\documentclass[runningheads]{llncs}

\usepackage[T1]{fontenc}
\usepackage{amsmath}
\usepackage{amsfonts}
\usepackage[table,xcdraw]{xcolor}
\usepackage{multirow}
\usepackage[squaren,Gray]{SIunits}
\def\doi#1{\href{https://doi.org/\detokenize{#1}}{\url{https://doi.org/\detokenize{#1}}}}

\usepackage{graphicx}
\usepackage{listings}
\begin{document}

\title{Multi-layer Aggregation as a key to feature-based OOD detection}
\titlerunning{Multi-layer Aggregation as a key to feature-based OOD detection}

\author{Benjamin Lambert \inst{1, 2} \and
Florence Forbes \inst{3} \and Senan Doyle \inst{2} \and Michel Dojat \inst{1}}
\authorrunning{B. Lambert et al.}

\institute{Univ. Grenoble Alpes, Inserm, U1216, Grenoble Institut Neurosciences, 38000, FR \and
Pixyl, Research and Development Laboratory, 38000 Grenoble, FR \and
Univ. Grenoble Alpes, Inria, CNRS, Grenoble INP, LJK, 38000 Grenoble, FR}
%

%
\maketitle              
\begin{abstract}
Deep Learning models are easily disturbed by variations in the input images that were not observed during the training stage, resulting in unpredictable predictions. Detecting such Out-of-Distribution (OOD) images is particularly crucial in the context of medical image analysis, where the range of possible abnormalities is extremely wide. Recently, a new category of methods has emerged, based on the analysis of the intermediate features of a trained model. These methods can be divided into 2 groups: \emph{single-layer} methods that consider the feature map obtained at a fixed, carefully chosen layer, and \emph{multi-layer} methods that consider the ensemble of the feature maps generated by the model. While promising, a proper comparison of these algorithms is still lacking. In this work, we compared various feature-based OOD detection methods on a large spectra  of OOD (20 types), representing approximately 7800 3D MRIs. Our experiments shed the light on two phenomenons. First,  \emph{multi-layer} methods consistently outperform \emph{single-layer} approaches, which tend to have inconsistent behaviour depending on the type of anomaly. Second, the OOD detection performance highly depends on the architecture of the underlying neural network. 
\keywords{Uncertainty \and Deep learning \and Anomaly Detection \and Medical images analysis }
\end{abstract}

\section{Introduction}
Out-of-distribution (OOD) images correspond to samples that are significantly different from the ones observed during training. Deep Learning (DL) models tend to behave inconsistently for this type of inputs, making OOD image detection crucial to avoid hidden model deficiencies. It is especially required in real-world automated pipelines, where input images may not be visually inspected before running the analysis. In the context of medical-images analysis, a large variety of phenomenons in the input images can impact a model and lead to unpredictable responses: noise, artifacts, variations in the imaging acquisition protocol and device, or pathological cases that were not included in the initial training dataset. Various methods were proposed for their detection, which can roughly be divided into two different categories \cite{berger2021confidence}: methods that build a model specifically dedicated to OOD detection; and methods that rely on the uncertainty or intermediate activations of a task-specific model (e.g image segmentation) to detect abnormal inputs. 

Within the first category, the most straightforward approach is to build a classifier to directly detect OOD images. For this, a Convolutional Neural Network (CNN) can be trained in a supervised manner, thus requiring the construction of an annotated dataset containing various types of real-world OOD \cite{bottani2022automatic}. Unsupervised Anomaly Detection (UAD) proposes to model the appearance of normal images by training an Auto-Encoder network (AE) to reconstruct in-distribution (ID) samples. At test-time, reconstruction is expected to be degraded for OOD samples, allowing for their detection \cite{gong2019memorizing}. 

Among the second category, uncertainty-based methods propose to detect OOD inputs directly from the outputs of an existing neural network. They rely on the hypothesis that the uncertainty of the deployed model should be high in the presence of a train-test mismatch, allowing its detection. A standard method consists of producing a set of diverse and plausible predictions for the same input image, with MC dropout \cite{gal2016dropout}, Deep Ensemble \cite{lakshminarayanan2017simple} or Test Time Augmentation \cite{wang2019aleatoric} being popular approaches. Uncertainty can then be estimated by computing the variance among the predictions. Alternatively, feature-based methods propose to analyse the intermediate activations of an existing model to detect OOD inputs \cite{postels2021practicality}. It is based on the assumption that the hidden activations of the model should be different for an ID image compared to an OOD image. A taxonomy of these feature-based methods is possible based on the number of layers used for OOD detection. \emph{Single-layer} methods only target one specific convolutional layer. In the context of medical image segmentation, popular choices are the end of the encoder \cite{gonzalez2022distance} or the penultimate convolutional layer \cite{karimi2020improving,diao2022unified}. \emph{Multi-layer} methods are an extension of the former that consider the entire set of convolutional layers in the trained model for OOD detection \cite{ccalli2022frodo}. Although gaining popularity, a proper comparison of these algorithms is still lacking. The contributions of our work are as follows:

\begin{itemize}
    \item We develop a large MRI segmentation benchmark comprising 20 different OOD datasets of various types and strengths, representing 7796 3D MRI volumes. We use this benchmark to compare 5 different feature-based OOD detectors as well as one uncertainty baseline.
    \item We adapt single-layer methods to a multi-layer fashion to demonstrate the potential performance gain that can be obtained with this enhancement. 

\end{itemize}

\begin{figure*}[!ht]
\centering
\includegraphics[width=0.95\textwidth]{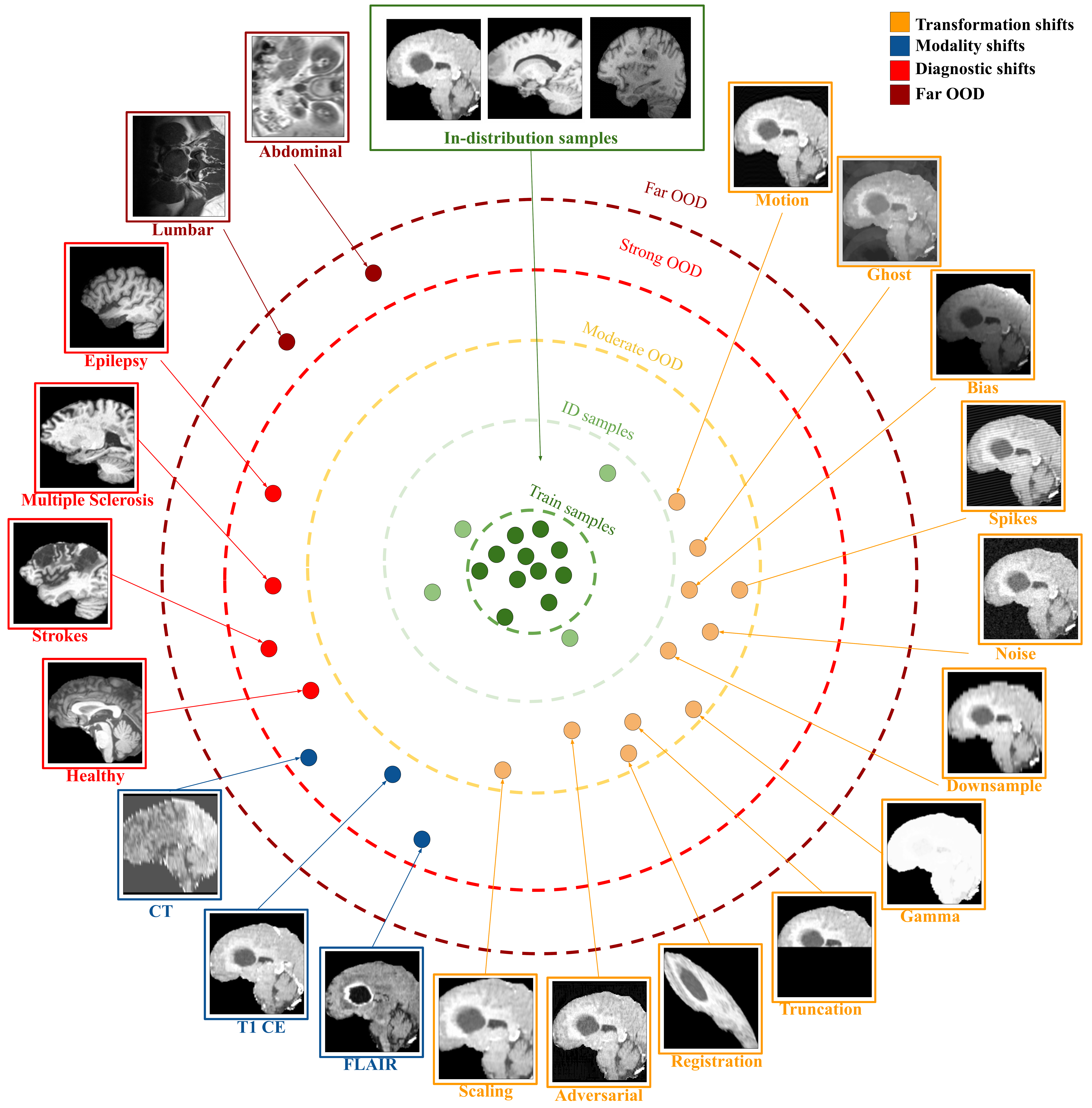}
\caption{Illustration of ID and OOD data used in the experiments.}\label{fig:ood}
\end{figure*}

\section{Compared methods}
\subsection{Feature-based methods}
Feature-based methods rely on a trained segmentation model and follow the same principle. First, a set of feature maps $F_i \in \mathbb{R}^{N\times H_i \times W_i \times D_i}$ is collected from a ID dataset for one convolution layer  $i$ (single-layer methods) or all convolution layers (multi-layer methods). Here, $N$ correponds to the number of convolutional filters in the $i$-th layer, and $H_i \times W_i \times D_i$ to the spatial dimensions of the feature map. Second, at inference time, a metric is computed to estimate the distance between the test features and the ID features to detect OOD samples. 

\paragraph{\textbf{Spectral signatures}} \cite{karimi2020improving} was proposed as a single-layer method focusing on the features obtained at the penultimate convolutional layer. Features are flattened to a 2D matrix $F_{2D} \in \mathbb{R}^{N\times HWD}$, and its singular values $S$ are calculated. The spectral signature is then taken as $\phi = \frac{log(S)}{\lVert log(S)\rVert_2}$. To detect OOD at test time, the distances $d_j$ between the signature of the test image $\phi_{test}^i$ and the signatures of a set of ID samples $\phi_{ID}^j$ is obtained by using the Euclidean distance. The final proposed OOD score corresponds to the minimum of the $d_j$ distances. 

\paragraph{\textbf{Prototypes}} \cite{diao2022unified} is a single-layer method that operates from the penultimate layer features and the segmentation masks predicted by the segmentation network. To obtain a prototype for a specific class and input image, features are multiplied with the binarized class mask, and average pooling is applied on the masked features. This yields to prototypes $P$ $ \in \mathbb{R}^{N \times C}$, $C$ being the number of segmented classes. An average ID prototype $P_{ID}$ is finally obtained by averaging the prototypes collected on the ID dataset. At test time, the OOD score is taken as the cosine dissimilarity between $P_{ID}$ and the test image prototype $P_j$. 

\paragraph{\textbf{The Mahalanobis Distance}} (MD) was recently investigated in 2 distinct studies for OOD detection in 3D medical images. In \cite{gonzalez2022distance}, authors focus on the features from the end of the encoder part of the network. They apply consecutive average pooling until the number of elements $M$ in the feature maps falls below a defined threshold of $1e4$, and then flatten it to obtain 1D vectors $z_i$ $\in \mathbb{R}^{M}$, for each ID image $i$. From these vectors, they compute the parameters of a multivariate Gaussian: the mean $\mu \in\mathbb{R}^{M}$ and covariance $\Sigma \in \mathbb{R}^{M\times M}$. At test time, the MD is computed given the fitted Gaussian and the test image feature representation. We refer to this single-layer method as \emph{MD Pool}. A similar approach is implemented in the multi-layer Free Rejection of Out-of-Distribution (FRODO) approach \cite{ccalli2022frodo}. This work differs in two ways: first, they directly compute the average of the feature map over the spatial dimensions ($H \times W \times D$) instead of applying average poolings. Second, they fit a multivariate Gaussian independently for each convolutional layer and compute the final OOD score as the average of each layer score. 

\paragraph{\textbf{One-class SVM}} (OCSVM) is an unsupervised algorithm for OOD detection that can be trained using only ID samples. It aims at finding the optimal boundary around the expected (ID) data. At test time, the distance to the boundary can be used as an OOD score, with ID sample being attributed with negative distances, and OOD samples with positive distances. Similar to \cite{wang2022layer}, we fit a OCSVM per convolution based on the averaged layer activations obtained from the ID images. At test time, each OCSVM produces a score, and the final OOD score is taken as the maximum of these scores.

\subsection{\textbf{Adapting single-layer methods to multi-layer methods}}
To assess the contribution of multi-layer aggregation to OOD detection, we propose to adapt single-layer methods (Spectrum, Prototypes and MD Pool) to multi-layer style. To achieve this, we replicate the OOD score computation step for each convolutional layer independently, yielding to \emph{layer-wise} scores $l_i$. As in FRODO, the final multi-layer score $L_{multi}$ is taken as the average of the scores of the individual layers: 

\begin{equation}
    L_{multi} = \frac{1}{N} \sum^N_{i=1} l_i
\end{equation}

\subsection{Uncertainty baseline}
To compare feature-based OOD detectors with a more traditional uncertainty methodology, we implement a MC dropout baseline \cite{gal2016dropout}. In MC dropout, the dropout layers of the segmentation model are kept activated at test-time and $N=20$ predictions are repeatedly sampled for each input image. The average voxel-wise variance among the MC samples is taken as uncertainty estimate at test time. 

\section{Material and Method}
\subsection{In-distribution Datasets}
Our work relies on the open-source BraTS 2021 dataset \cite{baid2021rsna} containing 1251 patients. The dataset initially includes four MRI sequences for each patient with four ground truth segmentation masks: the background, the necrotic tumor core, the edematous and the GD-enhancing tumor. We choose to focus on T1w sequences as this sequence is common and sufficient for experiment with multiple OOD settings. We also simplify the prediction task by focusing on the segmentation of the \emph{whole tumor core}, concatenation of all tumors sub-classes. The dataset is randomly split into a training fold (651), a calibration fold (200) used to fit the OOD detectors and a testing fold (400) (referred to as \textit{Test ID} in the following).
Additionally, we propose to include \emph{control} samples in our protocol, representing images that share the same properties than the training samples (same modality, organ and pathology), but that were acquired in a different imaging center. An effective model \emph{should} be able to generalize to these images and thus, the OOD detector should identify them as ID samples to prevent false alarms. We thus propose to use the LUMIERE glioblastoma dataset \cite{suter2022lumiere} as a \emph{Control} dataset, from which we select 74 T1-w pre-operative brain MRI. Figure \ref{fig:ood} illustrates the data used in the different experiments.

\subsection{Out-of-distribution Datasets}
Following the categorization of \cite{gonzalez2022distance}, we propose to investigate \emph{Transformation}, \emph{Diagnosis} and \emph{Far} OODs, as well as a new proposed setting, \emph{Modality} shifts.

\paragraph{\textbf{Transformation shifts.}} Finding real images with a controlled amount of artifacts to allow evaluation of OOD detection methods is difficult. We therefore generate realistic synthetic artifacted images from the set of \emph{Test ID} images \cite{fuchs2021practical,gonzalez2022distance}. We used the TorchIO Data Augmentation library [19] to generate \emph{Bias}, \emph{Motion}, \emph{Ghost}, \emph{Spikes}, \emph{Downsample}, \emph{Noise}, and \emph{Scaling} artifacts. We add a set of novel transformations: the \emph{Registration} that applies noise to the registration matrix to simulate an erroneous registration, the \emph{Gamma}  that applies extreme gamma modification to the image to mimick errors in the intensity normalization step, and the \emph{Truncation} that crops half of the brain. Finally, we also implement \emph{Adversarial} attacks, using the popular Fast Gradient Sign Method (FGSM) \cite{goodfellow2014explaining}. 

\paragraph{\textbf{Diagnosis shifts.}} DL segmentation models are usually trained with images showing a single 
 pathology (e.g brain tumor or strokes). However, once deployed, the model can be confronted with images exhibiting unseen anomalies, which can lead to incorrect predictions. To test OOD detection methods on this scenario, we use T1w brain MRI with various diseases: 170 subjects from the White Matter Hyperintensities (WMH) 2017 challenge \cite{kuijf2019standardized}, 655 subjects from the ATLAS-2 brain stroke dataset \cite{liew2022large} and 162 subjects from the EPISURG dataset \cite{perez2020simulation} containing epileptic subjects who underwent resective brain surgery. We also use 582 T1-w MRIs from healthy and young subjects from the IXI dataset \cite{ixi_dataset}.

\paragraph{\textbf{Modality shifts.}} Medical images are usually stored in DICOM formats, whose meta-data (headers) may be incorrectly filled \cite{gueld2002quality}. As a result, mismatches between the expected input modality (e.g T1w) and the test image modality (e.g CT or T2w) may be undetected. We construct 3 different \emph{Modality} shift OOD datasets. First, we use the FLAIR and T1ce sequences corresponding to the 400 test subjects. Second, we extract 437 brain CT-scans from the CQ500 dataset, exhibiting intracranial hemorrhage or cranial fractures.

\paragraph{\textbf{Far OOD}} corresponds to images that show little to no similarity with the ID samples. We use 2 non-brain T1w MRI datasets, respectively 80 abdominal MRI from the CHAOS dataset \cite{CHAOSdata2019} and 515 images from the Lumbar Spine MRI dataset \cite{natalia2018development}, as far OOD samples. 


\subsection{Influence of the segmentation model architecture}
Feature-based OOD detection methods rely on the hypothesis that the activations of the trained segmentation models are representative of the conformity of the input sample. To verify if this holds true for any segmentation model, we use the MONAI library \cite{cardoso2022monai} to train 6 different segmentation models: an Attention UNet (AttUNet) \cite{oktay2018attention}, a Residual UNet (ResUNet) \cite{kerfoot2018left}, a Dynamic UNet (DynUNet) \cite{isensee2018nnu}, a UNet++ \cite{zhou2018unet++}, a VNet \cite{milletari2016v} and a Transformer-based model, namely the UneTR \cite{hatamizadeh2022unetr}. All models are trained with the Dice loss \cite{milletari2016v}, instance normalization \cite{ulyanov2016instance}, a 3D dropout \cite{srivastava2014dropout} rate of $20\%$, and a batch size of 1, using the ADAM optimizer \cite{kingma2014adam} with a learning rate of $2e-4$. 

\setlength{\tabcolsep}{2pt}
\begin{table}[b!]
\footnotesize
\centering
\caption{Number of model parameters (in millions) as well as the average segmentation performance (Dice) on the \emph{Test ID} dataset, for each segmentation model.}\label{tab_dice}
\begin{tabular}{l|cccccc}
 & \multicolumn{1}{l}{AttUNet} & \multicolumn{1}{l}{ResUNet} & \multicolumn{1}{l}{VNet} & \multicolumn{1}{l}{UNet++} & \multicolumn{1}{l}{DynUNet}  & UNETR\\ \hline
Nb. parameters & 5.0 & 4.8 & 11.4 & 7.0 & 16.5 & 102 \\
\multicolumn{1}{c|}{Test ID Dice $\uparrow$} & \textbf{.83} & .81 & .81 & .81 & .82 & .76
\end{tabular}
\end{table}

\subsection{Evaluation Setting}
We cast OOD detection as a binary classification problem, where ID samples correspond to the positive class and OOD samples to the negative class. Each method produces a score for each OOD sample, which is compared with the scores obtained on the ID data in order to compute AUROC classification scores. We also report the segmentation performance using the Dice score, when the ground truth for brain tumors is available.

\setlength{\tabcolsep}{3.4pt}
\renewcommand{\arraystretch}{1}
\begin{table}[!t]
\fontsize{9pt}{9pt}
\centering
\caption{OOD detection performance (AUROC) for each dataset, obtained with the AttUNet segmentation model for the features and uncertainty approaches. We also report the number of samples in each dataset (N) and Dice score when the ground truth for brain tumors is available. S: single-layer. M: multi-layer.}
\begin{tabular}{lcc|cc|cc|cc|c|c|c}
 & & & \multicolumn{2}{c|}{Spectrum} & \multicolumn{2}{c|}{Prototypes} & \multicolumn{2}{c|}{MD Pool}& Frodo & SVM & MC \\ \hline
 Dataset & N & \begin{tabular}[c]{@{}c@{}}Dice\\ ($\uparrow$) \end{tabular} & S & M & S & M & S & M & M & M & -\\ \hline
 
 Test ID & 400 & .83 & - & - & - & - & - & - & - & - \\ 
 Control & 74  & .85 & 0.56 & 0.84 & 0.36 & 0.48 & 0.42 & 0.40 & 0.45 & 0.38 & 0.50 \\
Motion & 400 & .82 & 0.66 & \textbf{0.97} & 0.48 & 0.49 & 0.76 & 0.71 & 0.80 & 0.60 & 0.56 \\
Ghost & 400 & .80 & 0.66 & 0.57 & 0.46 & 0.48 & \textbf{0.88} & 0.85 & 0.85 & 0.60 & 0.63 \\
Bias & 400 & .78 & \textbf{1.00} & \textbf{1.00} & 0.86 & 0.82 & \textbf{1.00} & 0.98 & \textbf{1.00} & \textbf{1.00} & 0.77 \\  
Spikes & 400 & .79 & 0.90 & \textbf{1.00} & 0.69 & 0.82 & 0.86 & \textbf{1.00} & \textbf{1.00} & \textbf{1.00} & 0.63\\
Noise & 400 & .81 & 0.84 & \textbf{1.00} & 0.57 & 0.61 & 0.80 & \textbf{1.00} & \textbf{1.00} & \textbf{1.00} & 0.62 \\
Downsample & 400 & .82 & 0.69 & \textbf{0.97} & 0.48 & 0.50 & 0.55 & 0.55 & 0.73 & 0.60 & 0.53 \\
Gamma & 400 & .31 & \textbf{1.00} & \textbf{1.00} & 0.95 & 0.96 & 0.98 & 0.98 & \textbf{1.00} & \textbf{1.00} & 0.97 \\
Truncation & 400 & .61 & 0.99 & 0.99 & 0.86 & 0.83 & \textbf{1.00} & 0.99 & \textbf{1.00} & 0.99 & 0.63 \\
Registration & 400 & .01 & \textbf{1.00} & \textbf{1.00} & 0.93 & 0.88 & \textbf{1.00} & \textbf{1.00} & \textbf{1.00} & \textbf{1.00} & 0.84 \\
Adversarial & 400 & .58 & 0.77 & 0.69 & 0.42 & 0.50 & 0.89 & 0.95 & \textbf{1.00} & 0.97 & 0.80  \\
Scaling & 400 & .77 & 0.99 & \textbf{1.00} & 0.91 & 0.91 & 0.99 & 0.99 & \textbf{1.00} & \textbf{1.00} & 0.72 \\
\textbf{Transform} & \textbf{4400} & - & 0.86 & 0.92 & 0.70 & 0.71 & 0.89 & 0.92 & \textbf{0.95} & 0.89 & 0.70  \\
FLAIR & 400 & .10 & 0.99 & 0.99 & 0.40 & 0.91 & 0.99 & 0.97 & \textbf{1.00} & 0.99 & \textbf{1.00} \\
T1Ce & 400 & .69 & \textbf{0.97} & 0.96 & 0.80 & 0.78 & 0.92 & 0.92 & \textbf{0.97} & 0.90 & 0.90 \\
CT  & 437 & - & 0.99 & \textbf{1.00} & 0.36 & 0.89 & 0.99 & 0.98 & \textbf{1.00} & 0.99 & \textbf{1.00} \\
\textbf{Modality} & \textbf{1237} & - & 0.98 & 0.98 & 0.54  & 0.86 & 0.97 & 0.96 & \textbf{0.99} & 0.97 & 0.97\\  
Healthy & 577 & - & 0.55 & 0.99 & 0.89 & \textbf{1.00} & 0.12 & 0.57 & 0.97 & 0.98 & 0.92 \\
Strokes & 655 & - & 0.58 & 0.88 & 0.82 & \textbf{0.97} & 0.37 & 0.54 & 0.86 & 0.82 & 0.81\\
WMH & 170 & - & 0.73 & 0.96 & 0.89 & \textbf{0.98} & 0.30 & 0.54 & 0.96 & 0.93 & 0.76 \\
Epilepsy & 162 & - & 0.68 & 0.94 & 0.89 & \textbf{0.99} & 0.41 & 0.60 & 0.88 & 0.83 & 0.82 \\
\textbf{Diagnosis} & \textbf{1564} & - & 0.59 & 0.92 & 0.86 & \textbf{0.98} & 0.28 & 0.58 & 0.92 & 0.90 & 0.85 \\
Lumbar & 515 & - & \textbf{1.00} & \textbf{1.00} & 0.86 & 0.96 & \textbf{1.00} & \textbf{1.00} & \textbf{1.00} & \textbf{1.00} & \textbf{1.00} \\
Abdominal & 80 & - & \textbf{1.00} & \textbf{1.00} & 0.99 & \textbf{1.00} & \textbf{1.00} & \textbf{1.00} & \textbf{1.00} & \textbf{1.00} & \textbf{1.00} \\
\textbf{Far OOD} & \textbf{595} & - & \textbf{1.00} & \textbf{1.00} & 0.88 & 0.96 & \textbf{1.00} & \textbf{1.00} & \textbf{1.00} & \textbf{1.00} & \textbf{1.00} \\ 
\textbf{Overall} & \textbf{7796} & - & 0.84 & 0.93 & 0.72 & 0.81 & 0.79 & 0.86 & \textbf{0.95} & 0.91 & 0.80\\ 
\end{tabular}
\label{tab1}\end{table}

\begin{figure*}[htb!]
\centering
\includegraphics[width=\textwidth]{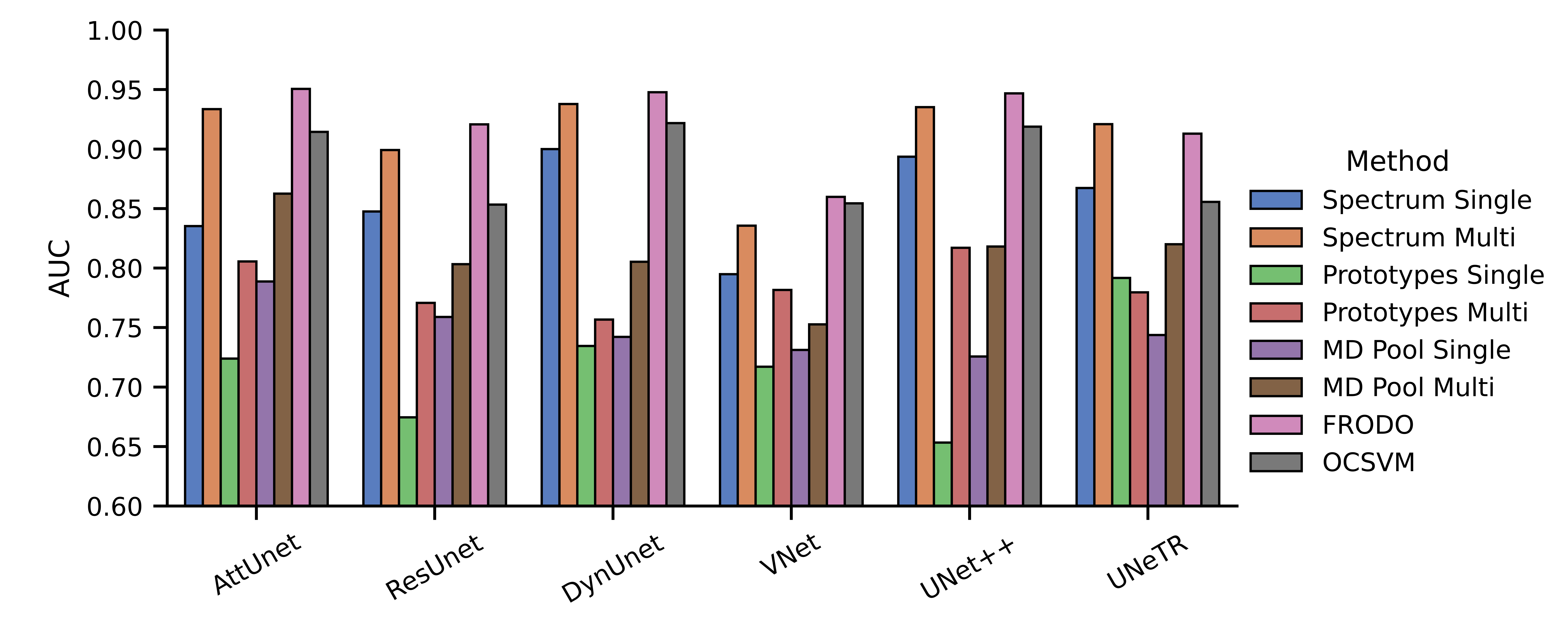}
\caption{\emph{Overall} OOD detection performance (AUROC score) of feature-based approaches with respect to the segmentation model architecture. }\label{fig:auc_wrt_archi}
\end{figure*}

\section{Results and Discussion}
The average segmentation performance of each tested segmentation backbone is presented in Table \ref{tab_dice}, with AttUnet being the best performer. OOD detection performances of each method are presented in Table \ref{tab1} using the AttUnet as backbone for feature and uncertainty-based methods. Finally, Figure \ref{fig:auc_wrt_archi} presents the \emph{Overall} OOD detection performance with respect to the neural architecture.

All OOD detectors achieve high detection accuracy on \emph{Far OOD} with mixed performances on other OOD types, showing that restricting to extreme OOD examples is insufficient to robustly validate a method. The best performer  is FRODO, achieving a perfect detection of non-conform inputs (AUROC=$1.00$) in 12 out of 20 settings, followed by the multi-layer implementation of Spectrum, and OCSVM. Overall, multi-layer methods outperform their single-layer version. For the AttUnet, this enhancement allows an increase of the \emph{Overall} AUROC score of $10.7\%$ for Spectrum, $12.5\%$ for Prototypes and $8.9\%$ for MD Pool. Single-layer methods exhibit variable performances depending on the OOD type, in accord with observations on 2D image classification   \cite{wang2022layer}. Finally, the MC dropout baseline has mixed performance in our benchmark, being outperformed by all multi-layer feature-based detectors. 

The rankings of OOD methods is roughly the same with different segmentation architecture (Figure \ref{fig:auc_wrt_archi}), with FRODO and Spectrum Multi-layer being the two top-performers. Interestingly, converting single-layer methods to multi-layer methods is beneficial for all architectures, with a gain on the AUROC for 6 out of 6 backbones for Spectrum and MD Pool, and 5 out of 6 backbones for Prototypes. However, the global \emph{Overall} OOD detection performance is variable depending on the segmentation architecture, e.g. FRODO achieves an AUROC score of $0.95$ or of $0.86$ when implemented on an AttUnet or VNet respectively. This indicates that certain popular medical image segmentation architectures are more prone to \emph{feature collapse} \cite{van2020uncertainty}, mapping OOD images to ID feature representations. Several strategies have been proposed in the context of 2D image classification to alleviate this issue, such as adding Gradient Penalty \cite{van2020uncertainty}, Lipschitz constraints \cite{liu2020simple}, or  a reconstruction term in the loss \cite{postels2021practicality}. These methods aims at enforcing a discriminative feature space for OOD detection, requiring changes in the training paradigm of the model, possibly resulting in sub-optimal predictive performances. To summarize, our main findings are: 
\begin{itemize}
    \item Feature-based methods monitoring the activation of \emph{all} convolution layers are more performant and robust than methods only targeting a single layer, whose performance is highly variable depending on the type of OOD. 
    \item The performance of these methods is dependent on the underlying segmentation architecture, with some of them being more prone to feature collapse, undermining the sensibility of OOD detection.
\end{itemize}


%
%

\bibliographystyle{splncs04}
\bibliography{biblio}

\end{document}